\def\year{2021}\relax
\documentclass[letterpaper]{article} 
\usepackage{aaai21}  
\usepackage{times}  
\usepackage{helvet} 
\usepackage{courier}  
\usepackage[hyphens]{url}  
\usepackage{graphicx} 
\usepackage{natbib}  
\usepackage{caption} 
\usepackage{amsmath}
 \usepackage{amssymb}
\usepackage{bbm}
\usepackage{dsfont}
\usepackage{footmisc}
\usepackage{booktabs}
\usepackage{multirow}

\usepackage{color}
\usepackage{tikz}
\usepackage{calc}
\usepackage{makecell}
\usepackage[switch]{lineno}  %
\usepackage{subfigure}
\usepackage{caption}

\makeatletter
\newcommand{\printfnsymbol}[1]{%
  \textsuperscript{\@fnsymbol{#1}}%
}
\makeatother

\title{Hamming OCR: A Locality Sensitive Hashing Neural Network \\for Scene Text Recognition}
\author{
    Bingcong Li\thanks{Equal Contribution.},  
    Xin Tang\printfnsymbol{1}, 
    Xianbiao Qi\printfnsymbol{1},
    Yihao Chen,
    Rong Xiao
    \\
}
\affiliations{

    Visual Computing Group, Ping An Property \& Casualty Insurance\\
    
    libingcong894@pingan.com.cn, \{tangxint, qixianbiao, xiaorong\}@gmail.com, o0o@o0oo0o.cc
}

\begin{document}
\maketitle

\begin{abstract}
Recently, inspired by Transformer, self-attention-based scene text recognition approaches have achieved outstanding performance. However, we find that the size of model expands rapidly with the lexicon increasing. Specifically, the number of parameters for softmax classification layer and output embedding layer are proportional to the vocabulary size. It hinders the development of a lightweight text recognition model especially applied for Chinese and multiple languages. Thus, we propose a lightweight scene text recognition model named \textit{Hamming OCR}. In this model, a novel Hamming classifier, which adopts locality sensitive hashing (LSH) algorithm to encode each character, is proposed to replace the softmax regression and the generated LSH code is directly employed to replace the output embedding. We also present a simplified transformer decoder to reduce the number of parameters by removing the feed-forward network and using cross-layer parameter sharing technique. 

Compared with traditional methods, the number of parameters in both classification and embedding layers is independent on the size of vocabulary, which significantly reduces the storage requirement without loss of accuracy. Experimental results on several datasets, including four public benchmaks and a Chinese text dataset synthesized by SynthText\footnote{https://github.com/JarveeLee/SynthText\_Chinese\_version} with more than \textit{20,000} characters, shows that Hamming OCR achieves competitive results.
\end{abstract}

\section{Introduction}
\label{sec_introduction}
Scene text recognition \cite{shi2016robust, shi2016end, cheng2017focusing, li2019show, lu2019master, chen2020text}, which aims at extracting text content from images, has attracted enormous attention from both the academy and industry due to its great commercial value in various real-world applications. 
 
With the development of sequence modeling, many text recognition models \cite{liu2016star, cheng2018aon, liu2018char} have achieved remarkable results. Generally, most competitive text recognition models have an encoder-decoder architecture, which maps each input sequence into an output sequence of variable length. Connectionist Temporal Classification (CTC) is applied for text recognition \cite{shi2016end, liu2016star, he2015reading, wang2017gated} to obtain the sequence of characters corresponding to the text image without character-level segmentation. \cite{shi2016end} integrates Recurrent Neural Network (RNN) with CTC to extract rich contextual information and make the decoding parallel and fast. However, these CTC-based methods are insufficient to deal with irregular text. 

\begin{figure}[t]
    \centering
    \includegraphics[scale=0.23]{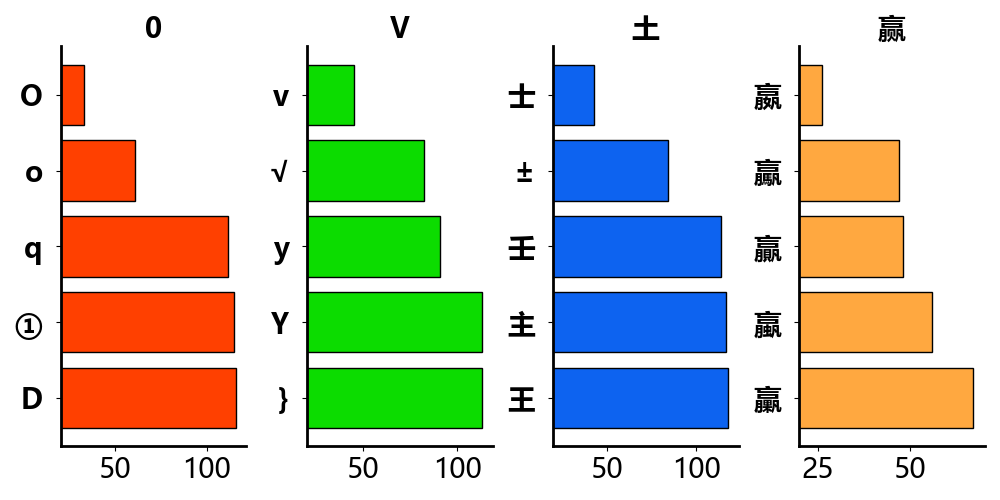}
    \caption{Hamming distances between characters. The vocabulary size is over $20,000$. Several characters are selected and their top $5$ nearest characters are shown. Note that the LSH code for each character contains $512$ bits.}
    \label{hamming_distance_fig}
\end{figure}

Recently, the attention mechanism, which is proposed to tackle machine translation in \cite{bahdanau2014neural}, is widely used to handle text recognition. For RNN-attention-based models \cite{lee2016recursive, cheng2017focusing, ghosh2017visual}, CNN and RNN are used as the encoder to extract contextual feature and another RNN combined with attention mechanism is employed as the decoder to align and decode character at each time step. A 2D attention mechanism, which can learn to select individual character features in the 2D space for decoding character, is proposed by \cite{li2019show}. \cite{wang2020decoupled} proposes a convolutional alignment module with UNet-style network architecture to address the misalignment issue. Inspired by the Transformer \cite{vaswani2017attention}, \cite{yang2020holistic, lu2019master} employ several transformer units to model the dependence among the sequence of output characters and align the visual features with the output character at each time-step.   

Many state-of-the-art works adopt heavy models. For instance, Master \cite{lu2019master} uses ResNet31 \cite{he2016deep} and several transfomer units. It is over 220MB. Obviously, it’s impractical to directly deploy these models on mobile system due to the immense storage overhead and computational cost. PaddleOCR\footnote{https://github.com/PaddlePaddle/PaddleOCR} provides a super lightweight CRNN \cite{shi2016end} using MobileNetV3 \cite{howard2019searching} to reduce model size from 31.8MB to 4.5MB.   

The softmax regression layer used in these methods, contains a projection matrix $W$ with size $d\times L$, where $d$ is the dimension of the character-level feature and $L$ is the length of vocabulary list. To reduce the storage cost, we need a small $L$ and small $d$. However, small $d$ hurts the model performance and small $L$ limits its application in many scenarios, such as Chinese and multi-language character recognition. For example, the storage cost of matrix $W$ is $39.1$MB when the vocabulary size is $20,000$ and $d$ is $512$. More over, in \cite{li2019show, lu2019master}, a matrix with similar size is used in the embedding layer to encode the output character. 

To address these issues, we propose a lightweight model, named Hamming OCR, to support scene text recognition with a large scale vocabulary. Hamming OCR is composed of four components: feature encoder, transformer decoder, Hamming embedding, and Hamming classifier, as shown in Figure \ref{fig:framework}. 

In the Hamming classifier, we use LSH \cite{gionis1999similarity} to map the output of a pre-trained model's feature layer to Hamming space. A majority voting mechanism is used to generate the representation code for each character. Due to the natural of LSH code, visual similar characters will be mapped codes with small Hamming distance, as shown in Figure \ref{hamming_distance_fig}. In the training stage, we employ a Hinge-loss to learn the optimal projection matrix for the target code.

Compare with the traditional one-hot encoding used in softmax, LSH code is multi-hot and more efficient when the vocabulary size is large. For example, given a vocabulary with  $20,000$ characters and a feature layer of $512$-dimension, the storage cost of the Hamming classifier is $1.22$MB, which is $32$ times smaller than softmax regression. Moreover, the LSH code is directly used in the output embedding, and the number of model parameters is further reduced.

Like \cite{lu2019master, yang2020holistic, yu2020towards}, we solely use transformer units to align and decode characters. To reduce the number of parameters, we remove the feed-forward component of the transformer unit and propose cross-layer parameter sharing for different transformer units. To deploy the model on mobile devices, we balance performance and model size, MobileNetV2 \cite{sandler2018mobilenetv2} is chosen as the feature encoder.

In summary, our contributions are summarized as follows:
\begin{itemize}
    \item We present a method to generate the LSH code to map each character to Hamming space. In this space, visual similar characters will have small Hamming distance. 
    \item We propose a novel Hamming classifier trained by Hinge-loss, which predicts output character using multi-hot LSH encoding instead of one-hot encoding. Using this strategy, the model's storage cost is significantly reduced when the vocabulary is large. 
    \item The LSH code is directly used in the output embedding module. It further reduces the computational cost and model size.
    \item We also simplify the transformer decoder architecture by removing the feed-forward module and cross-layer parameter sharing. 
    \item Hamming OCR delivers competitive results on several scene text datatsets, which is based on self-attention mechanism. More importantly, it can handle a large scale vocabulary and its size is very small. 
\end{itemize}

\section{Related Work}
\label{sec_relatedwork}
Given an input image $I$, the goal of scene text recognition is to produce a sequence $(y_1,y_2,...,y_T)$, where $y_t\in\{1,2,...,L\}$ is the character indicator, $L$ is the length of a predefined vocabulary list $V$. 

Generally, attention-based text recognition models \cite{li2019show, lu2019master, yu2020towards} have higher accuracy on irregular text datasets, which have an encoder-decoder architecture. Here, we mainly focus on attention-based models and divide a model into four basic components according to their role, including feature encoder, decoder, output embedding, and classifier. 

The first component is feature encoder which maps the input text image to a representation. In order to extract high-level visual features, ResNet is regard as the most popular CNN. For instance, \cite{hu2020gtc, yu2020towards, baek2019wrong} adopt ResNet50 as the feature encoder's backbone, and \cite{lu2019master, ghosh2017visual} use ResNet31 as the feature encoder. Meanwhile, many other types of ResNet \cite{li2019show, shi2018aster} also are utilized. To enlarge the feature context, RNN over feature sequence is adopted \cite{shi2018aster, shi2016end}. However, ResNet is inconvenient to deploy on the mobile system due to its storage requirement.

Decoder is the second component, which is used for sequence modeling. \cite{li2019show} adopts a 2-layer RNN with a 2D attention mechanism to decode the holistic feature into a sequence of characters. With transformer achieving success in natural language processing, \cite{lu2019master, yang2020holistic, yu2020towards} utilize self-attention module as decoder to learn character dependencies. They directly use the Transformer decoder which is composed of a masked self-attention mechanism to model relations between different characters of the output sequence, an attention module aligning character-level features from encoder with the output characters, and a feed-forward layer. However, the storage consuming of transformer decoder cannot be ignored. For instance, a typical feed-forward layer contains two projection matrices with size $2048\times 512$. 

Output embedding is widely used for sequence model. For text recognition, it encodes the output character of the previous time step as input for the decoder to decode the next character. Usually, learned embeddings \cite{lu2019master, vaswani2017attention} are employed to convert the input characters to vectors of dimension $d$. If we have $L$ characters for embedding, the size of the weight matrix of output embedding is $L\times d$. As the number of characters increases, the memory consumption of output embedding is unbearable for mobile devices. 

One of the most common classification approaches is softmax regression. Most text recognition methods directly use softmax regression to map the character-level features into probabilities over the vocabulary $V$. In the following subsection, we introduce softmax regression.   

\subsection{Softmax Regression for Classification}
Text recognition is a sequence prediction problem. At each time step, decoder extracts a character-level feature vector by use of attention mechanism. Then the character-level feature vector is mapped into the probability distribution over $V$ as
\begin{equation}
\label{distribution_eqn}
\begin{split}
\Pr(y|h, W) = \frac{\operatorname{exp}{(w_{y}^Th)}}{\sum_{j=1}^{L}\operatorname{exp}{(w_j^Th)}}, \\
\end{split}
\end{equation}
where $h\in\mathcal{F}$ denotes the character-level feature generated by decoder at each time step. $\mathcal{F}$ represents a $d$-dimensional character-level feature space.  $w_j\in\mathbb{R}^{d}$ denotes the $j$-th column of the weight $W\in\mathbb{R}^{d\times L}$. Then, classification decision is given by
\begin{equation}
\label{argmax_eqn}
\begin{split}
\hat{y} = \mathop{\arg\max}_{j\in\{1,2,...,L\}} \Pr(y=j|h, W).
\end{split}
\end{equation}
In the training process, to maximize the probability of the ground-truth sequence at each time step, hence, the cross-entropy loss $-\log(\Pr(y|h, W))$ is employed.  

\begin{figure*}
     \centering
     \begin{subfigure}
         \centering
         \includegraphics[width=.4\textwidth, height=.5\textwidth]{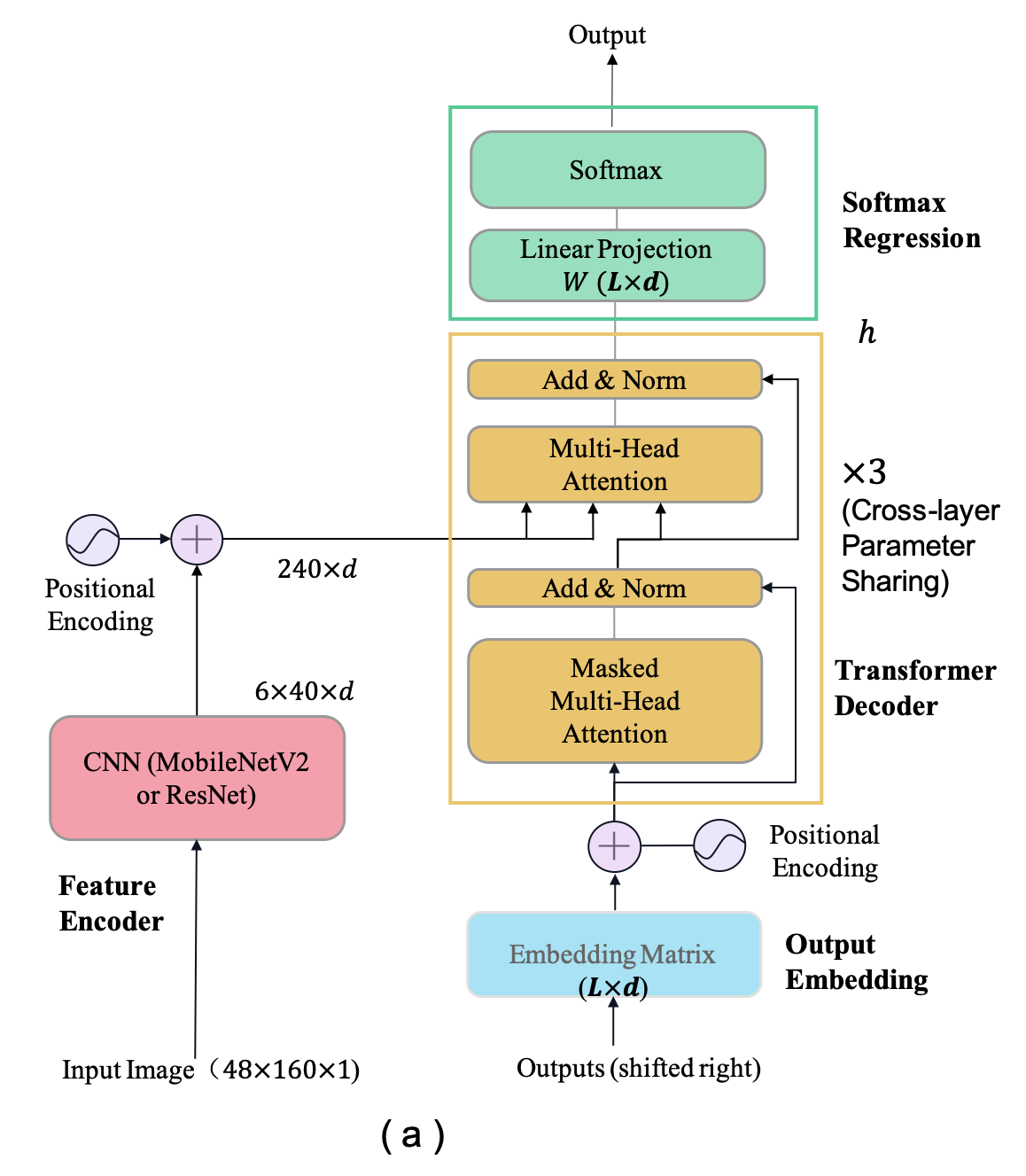}
     \end{subfigure}%
     \hfill
     \begin{subfigure}
         \centering
         \includegraphics[width=.4\textwidth, height=.5\textwidth]{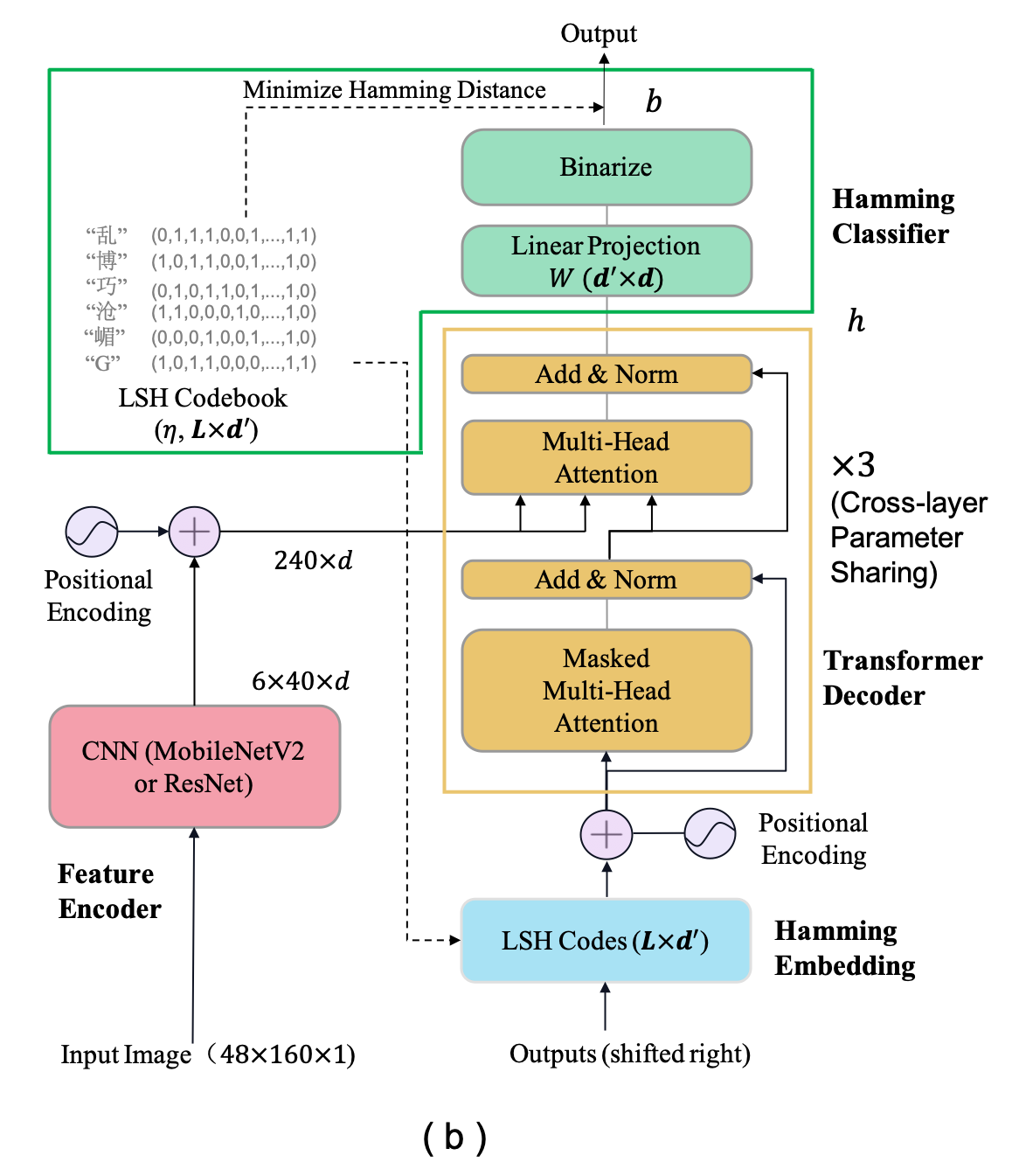}
     \end{subfigure}
        \caption{(a) The model architecture of the auxiliary model.  (b) The architecture of Hamming OCR model.}
        \label{fig:framework}
\end{figure*}

\section{Hamming OCR Model}
\label{sec_lightocr}
As shown in Figure \ref{fig:framework}(b), the Hamming OCR model contains four main components, feature encoder, transformer decoder, Hamming embedding, and Hamming classifier. In this section, we will first introduce the idea of LSH codebook for Hamming OCR model, then we will give the detail of each model components. 

\subsection{LSH Codebook for Hamming OCR Model}
Different from the traditional OCR models, we use multi-hot codes as the training target, and the model output is also directly used in the output embedding layer. According to the experimental result, which will be shown later in Table \ref{tab:code}, the design of the multi-hot codebook isn't trivial. Randomly generated codebook will hurt the model performance significantly. Therefore the Hamming OCR model cannot be trained directly without the LSH codebook.

To address this issue, we use an auxiliary model which replace the Hamming classifier with softmax regression, and Hamming embedding with the traditional embedding technique, which is shown in Figure \ref{fig:framework} (a). This model can be trained end to end directly. 

Using this model, any input text image is mapped into a sequence of character-level feature vectors $\{h_t\in\mathcal{F}\}_{t=1}^T$ for classification. Since the classifier isn't dependent on time-step $t$, we omit the time-step index and define $h_{i,j}\in\mathcal{F}$ to represent the $j$-th feature of $i$-th character class. Then we use LSH algorithm to project each feature vector $h_{i,j}$ to a $d'$-bits binary vector $b_{i,j}$ as follow
\begin{equation}
 \label{eqn:lsh}
     b_{i,j}^k = sgn(\psi_k^Th_{i,j}),
 \end{equation}
where $b_{i,j}^k$ is the $k$-th bit of the vector $b_{i,j}$, $\psi_k$ is the $k$-th vector of the random projection matrix $\Psi\in\mathbb{R}^{d\times d'}$. 

Since the softmax loss has the tendency to force the feature vectors from the same class to be close in the feature space, after the LSH mapping, the codes from the same character class will be close in the Hamming space as well. Based on this assumption, a majority vote algorithm is used to generate representation code for each character class.
\begin{equation}
\label{eqn:hamming_code}
    \eta_i^k=I(\sum_j{b_{i,j}^k}>n_i/2),
\end{equation}

\noindent where $\eta_i^k$ is the $k$-th bit of the LSH code $\eta_i$ for the $i$-th character class, $n_i$ is the number of feature vectors in the $i$-th class, and $I(.)$ is the indicator function which has the value 1 when the input is true and has the value 0 when the input is false. Finally, we get the codebook $\eta=[\eta_1,\eta_2,...,\eta_L]$ corresponding to the vocabulary $V$.

Theoretically, there is a chance that two character class has the same representation vector. However, when $d>256$, the chance of such conflict is small. Actually, even for the similar characters, The Hamming distance between the corresponding codes is not small. For example, visually, the character ``0'' looks like ``o''. Their LSH codes generated by our method also are similar to each other. From Figure \ref{hamming_distance_fig}, the Hamming distance between ``0'' and ``o'' is much smaller than the others, which is $33$. It means that the edit distance between their LSH code is $33$.

\subsection{Hamming Classifier}

For each input vector $h_{i,j}$, the Hamming classifier will output a binary vector of $d'$ bits, using the equation:

\begin{equation}
    b_{i,j}^k=\mathop{sgn}(w_k^Th_{i,j}),
    \label{eq:inference}
\end{equation}

\noindent where $b_{i,j}^k$ is the $k$-th bit of the output binary vector $b_{i,j}$, $w_k$ is the $k$-th column of the projection matrix $W$ used in the Hamming classifier. We hope that the output code $b_{i,j}$ should be close to the target code $\eta_i$ in the Hamming space. 

To achieve this goal, we propose to use the Hinge-loss \cite{cortes1995support} to train the whole model and find optimal $W$. The loss is defined as follow:
\begin{equation}
\label{loss_eqn}
\begin{split}
    \mathcal{L} = & \sum_{i,j,k}\{  \mathop{\max}\{0,\theta-w_k^Th_{i,j}\}\eta_i^k + \\
       & \mathop{\max}\{0,\theta+w_k^Th_{i,j}\}(1-\eta_i^k)\},
       \end{split}
\end{equation}

\noindent where $\theta$ is the margin, $w_k$ is the $k$-th project vector of $W$, $\eta_i^k$ is the $k$-th bit of the target code for the $i$-th class.

The Hamming classifier will map a feature vector into a point in a multi-dimension space. In each dimension of the space, the points from the same class will distribute in the same side of the axis. Due to the nature of Hinge-loss, these points are also be pushed away from the origin. It will reduce the chance of encoding error. 

In the inference stage, we use the Equation (\ref{eq:inference}) to calculate the binary vector $b$ for each feature vector $h$. However, due to the possible classifier error, the binary code $b$ may not be exist in the codebook $\eta$. To address this issue, we use:
\begin{equation}
\label{argmax_eqn}
\begin{split}
\hat{y} = \mathop{\arg\min}_{j\in\{1,2,...,L\}} \mathop{HammingDistance(b, \eta_j)},
\end{split}
\end{equation}
to generate the prediction $\hat{y}$.

\subsection{Hamming Embedding}
This component is used to encode the output character. Traditional methods try to learn a $d$-dimensional vector as embedding for each character. As the number of characters increase, the memory consumption of output embedding also increases proportionally. When there is a large vocabulary, the storage cost of the output embedding is heavy. To address this issue, we use the output codes as embedding of output characters as shown in Figure \ref{fig:framework}.

\subsection{Feature Encoder}
We choose the backbone of Master \cite{lu2019master} as our network's backbone, which integrates ResNet31 with GCNet \cite{cao2019gcnet}. Considering the model size and time consuming, for mobile applications, a lightweight MobileNetV2, which is presented in Appendix A, is used as our feature encoder.

\subsection{Transformer Decoder}
Attention mechanism can align relevant visual features to the corresponding output character, which aggregates information from the entire input sequence. Inspired by Transformer, the decoder of Transformer is applied to model sequence prediction in text recognition. A standard decoder of Transformer is composed of a stack of $N$ identical layers, each of which has three sub-layers. The first sub-layer is a masked self-attention network, the second one is a self-attention network and the third one is a feed-forward network. $Y^{\ell}\in\mathbb{R}^{T\times d}$ that is embedding of the output sequence is denoted as the input tensor of the $\ell$-th layer, then, these three sub-modules can be written as the following equations:
\begin{subequations}
\begin{align}
        & Y^{\ell}=\cup_{k=1}^{H}{\operatorname{Att(Y^{\ell}_QW^Q_{k,\ell},Y^{\ell}_KW^K_{k,\ell},Y^{\ell}_VW^V_{k,\ell})}}\cdot W^O_{\ell} \label{transformera} \\
       & Y^{\ell}=\cup_{k=1}^{H}{\operatorname{Att(Y^{\ell}U^Q_{k,\ell},XU^K_{k.\ell},XU^V_{k,\ell})}}\cdot U^O_{\ell} \label{transformerb} \\
      &  Y^{\ell} = \operatorname{ReLU(Y^{\ell}W_1^{FF}+b_1^{FF})}W_2^{FF}+b_2^{FF} \label{transformerc}
\end{align}
\end{subequations}
where $\operatorname{Att}(\cdot)$ is scaled dot-product attention \cite{vaswani2017attention}, $H$ is the number of heads and $\cup_{k=1}^{H}$ denotes a concatenated operator. In Equation (\ref{transformera}), $Y^{\ell}_Q,Y^{\ell}_K,Y^{\ell}_V$ are masked $Y^{\ell}$ to prevent a given position from incorporating information about future output positions. $X$ in Equation (\ref{transformerb}) is extracted by the feature encoder. Note that residual connections and layer normalization are used for each sub-layer, which don't show in Equation (\ref{transformera}), (\ref{transformerb}) and (\ref{transformerc}).

For the $\ell$-th layer, $\{W_{k,\ell}^Q,W_{k,\ell}^K,W_{k,\ell}^V\}_{k=1}^{H}\in\mathbb{R}^{d\times \frac{d}{H}}$, and $W_{\ell}^{O}\in\mathbb{R}^{d\times d}$ are parameters corresponding to masked multi-head self-attention sub-layer. $\{U_{k,\ell}^Q,U_{k,\ell}^K,U_{k,\ell}^V\}_{k=1}^{H}\in\mathbb{R}^{d\times\frac{d}{H}}$, and $U_{\ell}^{O}\in\mathbb{R}^{d\times d}$ are parameters of multi-head self-attention modules. $W_1^{FF}\in\mathbb{R}^{d\times 2048}, W_2^{FF}\in\mathbb{R}^{2048\times d}$, and $b_1^{FF}\in\mathbb{R}^{2048},b_2^{FF}\in\mathbb{R}^{d}$ are parameters of feed-forward sub-layer. There are many parameters for one layer of transformer decoder.

\textit{Removing the feed-forward network and cross-layer parameter sharing} techniques are employed in our transformer decoder to reduce the storage requirement, which stacks three transformer decoder layers. From Equation (\ref{transformerc}), we find that the main function of a feed-forward network with two linear transformations is to learn a projection. However, in Equation (\ref{transformera}) and (\ref{transformerb}), each of them contains a projection matrix ($W_{\ell}^O,U_{\ell}^O)$ which has similar function with feed-forward network. More importantly, feed-forward network has many parameters. So, we remove the feed-forward network (Equation (\ref{transformerc})) in each layer of transformer decoder. For one layer of our transformer decoder, it contains two sub-layers. The first sub-layer is a masked self-attention network and another one is a self-attention network as is shown in Figure \ref{fig:framework}. Like \cite{lan2019albert}, cross-layer parameter sharing technique is used for all the three layers of our transformer decoder. That means, the parameters in Equation (\ref{transformera}) and (\ref{transformerb}) keep the same values for different layers.

\begin{table}[!h]
\centering
\caption{Model size of Hamming OCR under different settings.}
\resizebox{0.46\textwidth}{!}{%
\begin{tabular}{p{4.3cm}|p{1cm}<{\centering}p{1cm}<{\centering}p{1cm}<{\centering}p{1cm}<{\centering}p{1cm}<{\centering}p{1cm}<{\centering}p{1cm}<{\centering}}
\cmidrule[1.5pt]{1-8}
& \multicolumn{7}{c}{Hamming OCR}                                                \\ 
\hline
MobileNetV2        & $\times$ & \checkmark   & \checkmark   &  \checkmark   & \checkmark & \checkmark & \checkmark  \\ 
Hamming Classifier        & $\times$ &  & \checkmark & \checkmark & \checkmark & \checkmark & \checkmark \\
Hamming Embedding             & $\times$ &                           &  & \checkmark & \checkmark & \checkmark & \checkmark \\
No Feed-Forward                   & $\times$ &                           &                           &   & \checkmark & \checkmark & \checkmark  \\
Cross-layer Parameter Sharing & $\times$ &                           &                           &                           &  & \checkmark & \checkmark \\
Half-Precision Float (FP16)        & $\times$ &                           &                           &                           &                           &     & \checkmark \\ 
\cmidrule[1.5pt]{1-8}
Model Size  & 305.8Mb   &55.6Mb &36.7Mb &16.6Mb &10.6Mb &6.6Mb  &3.9Mb                      \\ 
\cmidrule[1.5pt]{1-8}
\end{tabular}%
}
\label{memory_consumption_tab}
\end{table}

\subsection{Parameter Analysis}
In this subsection, we analyze the storage requirement of different models under different settings. Our baseline model has similar with \cite{lu2019master}, which uses a RestNet31 with GCNet as CNN's backbone and three-layer transformer decoder. We choose a vocabulary with $20,948$ characters. Then, we compute that the baseline model requires 305.8Mb storage resource. When MobileNetV2 is used to replace ResNet31, the model size sharply reduces to 55.6Mb. Different from the baseline model, Hamming OCR adopts many techniques to reduce the model size including Hamming Classifier, Hamming Embedding, removing feed-forward network of transformer decoder, and cross-layer parameter sharing. Table \ref{memory_consumption_tab} shows the model size for each model. Our MobileNetV2-based Hamming OCR only costs $6.6$Mb. Additionally, when we use 16-bit floating point to represent the weights of Hamming OCR, the most lightweight Hamming OCR is 3.9Mb.

\section{Experiments}
\label{tab:experiments}
We conduct extensive experiments to verify the effectiveness of our Hamming OCR. First, several public standard datasets are employed. However, these public datasets only contain English words and the number of characters to be recognized is relatively small. In order to further validate the capability of Hamming OCR, a dataset named as GBK21K is newly generated.

\subsection{Datasets}

\textbf{IIIT 5K-Words (IIIT5K)} \cite{mishra2012top} contains $3,000$ cropped scene text images for testing.

\noindent\textbf{Street View Text (SVT)} \cite{wang2011end} consists of $257$ training images and $647$ testing images, which is collected from Google Street Image.

\noindent\textbf{SVT-Perspective (SVTP)} \cite{quy2013recognizing} consists of $645$ cropped images. Many images have perspective distortions.

\noindent \textbf{CUTE80 (CUTE)} contains $288$ text patches cropped from natural scene images for curved text recognition.

\noindent \textbf{MJSynth (MJ)}  \cite{jaderberg2014synthetic} consists of $9$ millions image instances, which is randomly generated based on $90$k English words.

\begin{figure}[!h]
    \centering
    \includegraphics[scale=0.5]{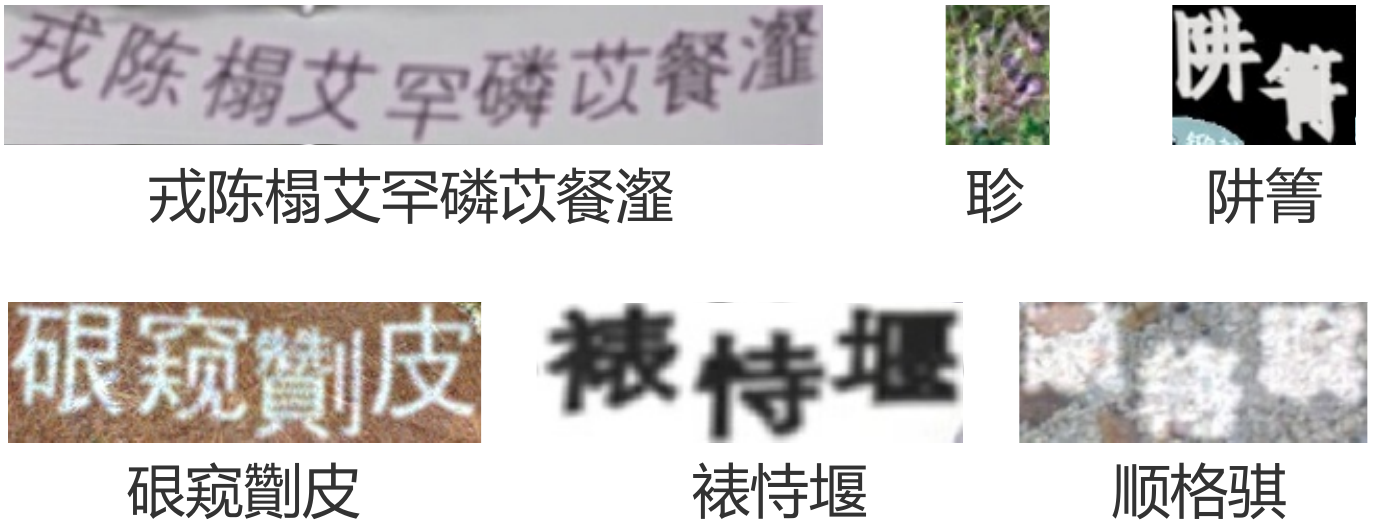}
    \caption{Examples of text in GBK21K. Several cropped images and corresponding ground truth texts are shown.}
    \label{fig:GBK21k}
\end{figure}

\noindent \textbf{GBK21K} is generated with the engine in \cite{gupta2016synthetic}, which includes $3$ million text patches for training and $30$k cropped images for testing. The text of each cropped image is generated by randomly selecting several characters from a vocabulary collected in advance. This vocabulary is a subset of the GBK Chinese characters \footnote{https://en.wikipedia.org/wiki/GBK\_(character\_encoding)} and contains $20,948$ characters including Chinese, English and numeric characters. The dataset is challenging due to low-resolution, clustered background, different fonts, and various illumination. Figure
\ref{fig:GBK21k} illustrates several cropped images and ground-truth texts. 

\subsection{Training Strategy}
The training of Hamming OCR includes two stages. In the first stage, we train a auxiliary model which uses ordinary output embedding and softmax regression. Cross entropy loss is used in this stage. With the trained model, we generate the codebook for all characters. In the second stage, we replace the output embedding with the Hamming embedding and the softmax regression with the Hamming classifier. The parameters of the backbone and transformer decoder are initially loaded from the model trained in the first stage. 
We retrain the model using Hinge-loss.

We implement our Hamming OCR with PyTorch and run all experments on NVIDIA Tesla V100 GPUs with 16GB memory. The batch size on each GPU is $160$, with $8$ GPUs in total. All input images are padded and resized to $48\times 160$. For standard benchmarks, without any data augmentation, we directly use synthetic data MJSynth (7.2M) as our training data. There are $62$ characters to be recognized. For the GBK21K dataset, we generate $3$M cropped images for training and evaluate the performance on $30$K text images. GBK21k is used for Chinese text recognition and contains $20,948$ symbol classes. We adopt the Adam optimizer, and the following hyper-parameters are used: the initial learning rate of $0.001$ and the decay rate of $0.5$.

\begin{table*}[!t]
\centering
\caption{Performance and model size (Mb) comparison on several public benchmarks and a Chinese text dataset. Master\textsuperscript{*} represents the model trained on an union of MJSynth (MJ), SynthText \cite{gupta2016synthetic} (ST) and SynthAdd \cite{gupta2016synthetic} (SA). PaddleOCR$\ddagger$ and PaddleOCR$\dagger$ represent the officially supplied model and the model trained by ourselves with the official code respectively. ``HC'' and ``HE'' mean Hamming classifier and Hamming embedding are used respectively, ``NoFFN'' and ``PS'' are corresponding to removing feed-forward network and cross-layer parameter sharing techniques for transformer decoder. The 
index ``Ratio (\%)'' represents the ratio of model size of Hamming classifier and output embedding in the whole model.}
\label{tab:performance}
\resizebox{\textwidth}{!}{%
\begin{tabular}{|m{5.4cm}||p{0.8cm}<{\centering}|m{1.3cm}|m{1cm}|m{1cm}|m{1cm}|m{1cm}||m{1.3cm}|m{1cm}|m{1cm}|}
\hline
\multicolumn{1}{|c||}{\multirow{2}{*}{\textbf{Methods}}} & \multicolumn{6}{c||}{\textbf{Public Benchmarks}} & \multicolumn{3}{c|}{\textbf{Chinese Text Recognition}}       \\ \cline{2-10} 
\multicolumn{1}{|c||}{}                                  & \multicolumn{1}{c|}{Training Data} & \multicolumn{1}{c|}{Mode Size} & \multicolumn{1}{c|}{IIIT5K} & \multicolumn{1}{c|}{SVT} & \multicolumn{1}{c|}{SVTP} & \multicolumn{1}{c||}{CUTE} & \multicolumn{1}{c|}{Mode Size} & \multicolumn{1}{c|}{Ratio ($\%$)} & \multicolumn{1}{c|}{GBK21K} \\ \hline
\hline
RARE \cite{shi2016robust} & MJ &$-$ &$81.9$ &$81.9$ &$71.8$ &$59.2$ &$-$ &$-$ &$-$ \\
Yang et al. \cite{yang2017learning} & MJ &$-$ &$-$ &$-$ &$75.8$ &$69.3$ &$-$ &$-$  &$-$ \\
R\textsuperscript{2}AM \cite{lee2016recursive} &MJ &$-$ &$78.4$ &$80.7$ &$-$ &$-$ &$-$ &$-$  &$-$\\
CRNN \cite{shi2016end}&MJ &$31.8$ &$78.2$ &$80.8$ &$-$ &$-$ &$72.7$ &$56.3$ &$-$                            \\
CRNN (VGG) \cite{shi2018aster} &MJ &$-$ &$81.2$ &$82.7$ &$-$ &$-$ &$-$ &$-$  &$-$                \\

ASTER-A \cite{shi2018aster} &MJ &$-$ &$81.7$ &$80.2$ &$73.2$ &$63.9$ &$-$ &$-$ &$-$                         \\
ASTER-B \cite{shi2018aster} &MJ &$80.4$ &$83.2$&$81.6$&$75.4$&$67.4$&$161.9$& $49.7$ & $-$                       \\
PaddleOCR$\ddagger$ (MobileNetV3, large) &MJ+ST &$-$ &$83.7$ &$84.1$ &$71$ &$62.2$&$-$& $-$  & $-$                            \\
ASTER \cite{shi2018aster} & MJ+ST&$-$ &$93.4$&$89.5$&$78.5$&$79.5$&$-$& $-$ & $-$                      \\
SAR \cite{li2019show}  & MJ+ST+SA & $-$ & $91.5$ & $84.5$ & $76.4$ & $83.3$  & $-$&$-$ & $-$  \\
Master\textsuperscript{*} \cite{lu2019master}  & MJ+ST+SA & $223.9$ & $95.0$ & $91.8$ & $84.5$ & $87.5$  & $305.9$  &  $13.4$  & $-$  \\
\hline \hline
Master (ResNet31)  & MJ& $223.9$ & $85.4$ & $85.3$ & $74.1$ & $\textbf{69.8}$  &$305.9$  & $13.4$  &  $82.3$  \\
HC (ResNet31) & MJ &$224.7$ &$85.7$ &$\textbf{85.6}$ &$74.1$ &$69.1$ &$267.0$ &$16.3$ &$82.4$         \\
HC+HE (ResNet31) & MJ& $224.6$ &$\textbf{85.9}$ &$\textbf{85.6}$ &$\textbf{74.6}$ &$68.4$ &$226.9$ &$1.4$ &$82.5$ \\
NoFFN+PS (ResNet31)& MJ &$184.0$ &$84.8$ &$83.9$ &$73.2$ &$\textbf{69.8}$ &$266.1$ &$30.9$ &$82.1$ \\
Hamming OCR (ResNet31) & MJ &$184.8$ &$85.0$ &$84.5$ &$73.0$ &$69.1$ &$187.1$ &$1.2$ &$\textbf{82.8}$ \\
\hline \hline
PaddleOCR$\dagger$ (MobileNetV3, small) & MJ &$2.7$ &$79.5$& $79.8$&$63.7$&$52.4$&$9.4$ & $82.7$  & $48.3$  \\
PaddleOCR$\dagger$  (MobileNetV3, large) & MJ &$5.3$ &$81.6$& $80.7$&$68.2$&$58.7$&$11.7$ &$66.4$  & $66.5$  \\
Hamming OCR (MobileNetV2) &MJ &$4.6$ &$82.6$ &$83.3$ &$68.8$ &$61.1$ &$\textbf{6.6}$ &$27.1$ &$\textbf{71.2}$ \\
\hline
\end{tabular}
}
\end{table*}

\subsection{Recognition Performance Evaluation}
The recognition accuracies of different methods on five datasets, including regular (IIIT5K, SVT), irregular (SVTP, CUTE) and a Chinese text (GBK21K), are shown in Table \ref{tab:performance}. In Hamming OCR model, four techniques are used including Hamming classifier, Hamming embedding, removing feed-forward network and cross-layer parameter sharing. To compare the effect of different techniques, we design three models based on the Master model by: replacing the softmax regression with Hamming classifier, namely HC; replacing the softmax regression and output embedding with Hamming classifier and Hamming embedding respectively, namely HC+HE; and employing removing feed-forward network and cross-layer parameter sharing technique, namely NoFFN+PS.

Firstly, we analyze the performance of different methods on the four standard benchmarks. Compared with previous methods with the same training data (MJ), Hamming OCR achieves competitive results. Especially, HC+HE has the similar performance as Master in most situations, which verifies the effectiveness of the proposed Hamming classifier and Hamming embedding. When we use removing feed-forward network and cross-layer parameter sharing techniques in transformer decoder module, the results of NoFFN+PS are slightly lower than Master. Due to the lightweight CNN backbone, the results of Hamming OCR (MobileNetV2) significantly decrease. However, Hamming OCR (MobileNetV2) consistently outperforms PaddleOCR in comparison with the same training data. For instance, the accuracy of Hamming OCR model on IIIT5K is higher than the larger PaddleOCR model by one percentage, while its model size is smaller. Aster has outstanding results on irregular datasets due to its rectification module.  

Secondly, On the GBK21K datatset, compared with Master, Hamming OCR not only obtains competitive performance but also sharply reduces the storage cost. The results show that Hamming OCR can handle such challenging Chinese recognition task. For the lightweight models, our Hamming OCR (MobileNetV2) outperforms PaddleOCR. Hamming OCR (MobileNetV2) includes a backbone of 2.3Mb, a simplified transformer decoder of 2.5Mb, a projection matrix of 0.5Mb and a codebook of 1.3Mb, while the PaddleOCR (MobileNetV3, large) includes a backbone of 3.0Mb, an LSTM decoder of 0.9Mb and a softmax classifier of 7.8Mb. It can been seen that softmax regression take a large portion of the storage cost of PaddleOCR.

In Hamming OCR, the projection matrix $\Psi$ used to generate the LSH codebook is randomly initialized. When we directly evaluate the performance of Hamming OCR with this initial projection matrix, its results are $50.9\%$ on IIIT5K, $48.5\%$ on SVT, $32.7\%$ on SVTP, $32.3\%$ on CUTE. Therefore, we confirm that retraining the model with the Hinge-loss is very vital.

\subsection{Ablation Study}

\subsubsection{Factorized Embedding Vs LSH Code.}
From a practical perspective, Chinese or multi-language text recognition usually require the length of the vocabulary list $L$ to be large, then increasing $L$ increases the size of the embedding matrix ($d\times L$) from output embedding and the projection matrix ($d\times L$) from softmax regression. As ALBERT \cite{lan2019albert}, we use a factorized embedding technique, which decomposes a matrix into two smaller matrices, to reduce the parameters. We take the linear transformation $W\in\mathbb{R}^{d\times L}$ of softmax regression as an example. It can be decomposed into $W_1\in\mathbb{R}^{d\times p}$ and $W_2\in\mathbb{R}^{p\times L}$. By using this decomposition, we reduce the parameters $W$ from $\mathcal{O}(d\times L)$ to $ \mathcal{O}(d\times p+p\times L)$. This parameter reduction is significant when $d>>p$.
 On the other hand, we present the LSH code can significantly reduce the parameters in classifier and output embedding modules. Thus, in this part, we take the experiments on GBK21K to evaluate the performance of factorized embedding technique and our Hamming classifier.
 
\begin{figure}[!h]
    \centering
    \includegraphics[scale=0.90]{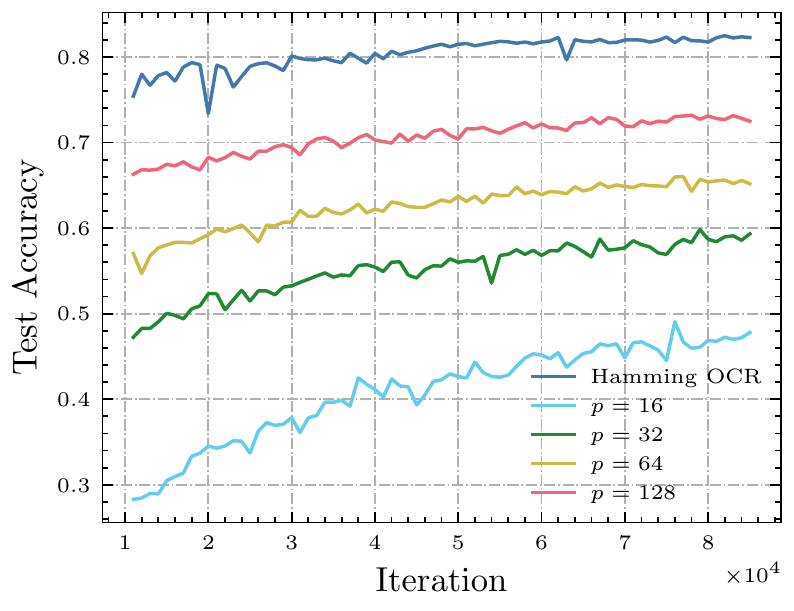}
    \caption{Performance comparison on GBK21K with different training steps. The performance of factorized embedding technique with $p=16,32,64,128$ are demonstrated.}
    \label{fig:embedding}
\end{figure}

From Figure \ref{fig:embedding}, we can see that as the value of $p$ increases, the recognition accuracy of the model based on factorized embedding technique is also increasing. Compared with factorized embedding technique, Hamming OCR performs better and its size is smaller.

\subsubsection{Random Code Vs LSH Code.}
In our Hamming OCR, we obtain LSH code by linear projection and thresholding. To verify the effectiveness of LSH code, we compare its performance with a random code, which is simply generated by a random method. To generate each element of the random code for a character, we randomly choose 0 or 1 with equal probability. To be fair, we adopt HC as the baseline model. Table \ref{tab:code} illustrates the recognition results on datasets. We can see that the performance of random code on the four standard benchmarks is competitive. For example, the recognition accuracy of LSH code is $85.6\%$ while random code can reach $84.5\%$ on the SVT dataset. However, for the GBK21K dataset, model with random code does not converge at all and can't recognize any text image correctly. As we know, the lexicon of SVT is very small, the random code can represent each character very well. However, there are over $20,000$ characters in GBK21K, it's possible that the representation ability of random code is insufficient. Different from random code, LSH code can reach $82.4\%$ recognition rate on GBK21K dataset.

\begin{table}[!h]
\centering
\caption{Performance comparison of baseline model with LSH code and random code on GBK21K. * means the model does not converge.}
\label{tab:code}
\resizebox{0.45\textwidth}{!}{%
\begin{tabular}{cccccc}
\toprule
 \textbf{Methods}        & \textbf{IIIT5K} & \textbf{SVT} & \textbf{SVTP} & \textbf{CUTE} & \textbf{GBK21K} \\ \midrule
Random Code                & $81.53$ &$84.54$ & $73.18$ & $68.06$ & $0.00^*$  \\ 
LSH Code              & $85.70$ &$85.63$ & $74.11$ & $69.10$ & $82.41$  \\ 
\bottomrule
\end{tabular}%
}
\end{table}

\subsubsection{Influence of The Length of LSH Code.}
Table \ref{tab:length} demonstrates the variation of recognition rate along with the length of LSH code. Obviously, LSH code with $512$-dimension performs very well. When the length of LSH code is less than $512$, the performance of Hamming OCR has declined. As the length of LSH code is larger than 512, the performance almost saturates. However, its negative impact is that the model size obviously becomes larger.

\begin{table}[!h]
\centering
\caption{Performance with different lengths of  LSH code on GBK21K.}
\label{tab:length}
\begin{tabular}{|c|c|c|c|c|}
\hline
Length (LSH Code) & $256$   & $512$   & $1024$  & $2048$  \\ 
\hline
GBK21K (accuracy)  & $81.86$ & $82.39$ & $82.26$ & $82.31$ \\ 
\hline
\end{tabular}%
\end{table}

\section{Conclusion}
\label{sec_conclusion}
In this paper, we present a lightweight and effective Hamming OCR method for scene text recognition. Most scene text recognition approaches adopt learnable output embedding and softmax regression for decoding the output characters and classification. Each of them has a parameter matrix, the size of which depends on the number of characters. when we need deploy a model with a large scale lexicon on mobile devices, these two matrices are encumbrances due to their massive storage requirements. Therefore, we present a method to generate a codebook for all characters and develop a Hamming classifier for classification. We further use the LSH code of each character as its embedding. As we know, this work is the first attempt to use LSH code as the representation vector of each class for classification and embedding. To reduce the parameters of a standard decoder of transformer, we introduce a simplified transformer decoder by removing the feed-forward network and using cross-layer parameter sharing technique. We combine Hamming classifier, Hamming embedding, and simplified transformer decoder into Hamming OCR. Hamming OCR achieve competitive results on on several standard benchmarks and one our own dataset.

\bibliography{ocr}
\end{document}